\documentclass[sigconf]{acmart}
\AtBeginDocument{%
	\providecommand\BibTeX{{%
			\normalfont B\kern-0.5em{\scshape i\kern-0.25em b}\kern-0.8em\TeX}}}

\copyrightyear{2020}
\acmYear{2020}
\setcopyright{acmcopyright}\acmConference[CIKM '20]{Proceedings of the 29th ACM International Conference on Information and Knowledge Management}{October 19--23, 2020}{Virtual Event, Ireland}
\acmBooktitle{Proceedings of the 29th ACM International Conference on Information and Knowledge Management (CIKM '20), October 19--23, 2020, Virtual Event, Ireland}
\acmPrice{15.00}
\acmDOI{10.1145/3340531.3411858}
\acmISBN{978-1-4503-6859-9/20/10}

\usepackage{soul}
\usepackage{url}
\usepackage[utf8]{inputenc}
\usepackage{graphicx}
\usepackage{amsmath}
\usepackage{amsthm}
\usepackage{booktabs}
\usepackage{algorithm}
\usepackage{algorithmic}
\usepackage{bm}
\usepackage{tabu}
\usepackage{diagbox}
\usepackage{mathrsfs}
\usepackage{amsfonts}
\usepackage{multirow, multicol}
\usepackage{CJKutf8}
\usepackage{color}
\usepackage{subfigure}

\newcommand{\figref}[1]{Figure \ref{#1}}

\newcommand{\tabref}[1]{Table \ref{#1}}
\newcommand{\secref}[1]{Section \ref{#1}}

\begin{document}
\fancyhead{}
	
\title{MICK: A Meta-Learning Framework for Few-shot Relation Classification with Small Training Data}

\author{Xiaoqing Geng}
\email{gxq961127@sjtu.edu.cn}
\affiliation{%
	\institution{Shanghai Jiao Tong University}
}
\author{Xiwen Chen}
\email{victoria-x@sjtu.edu.cn}
\affiliation{%
	\institution{University of Michigan-Shanghai Jiao Tong University Joint Institute}
}
\author{Kenny Q. Zhu}
\authornote{Corresponding Author.}
\email{kzhu@cs.sjtu.edu.cn}
\affiliation{%
	\institution{Shanghai Jiao Tong University}
}
\author{Libin Shen}
\email{libin@leyantech.com}
\affiliation{%
	\institution{Leyan Tech}
}
\author{Yinggong Zhao}
\email{ygzhao@leyantech.com}
\affiliation{%
	\institution{Leyan Tech}
}

\begin{CJK}{UTF8}{gkai}
	
\begin{abstract}
Few-shot relation classification seeks to classify incoming query instances after meeting 
only few support instances. 
This ability is gained by training with large amount of in-domain annotated data. 
In this paper, we tackle an even harder problem by further limiting the amount of data available
at training time. 
We propose a few-shot learning framework for 
relation classification,
which is particularly powerful when the training data is very small.
In this framework, models not only strive to classify query instances, but also seek 
underlying knowledge about the support instances
to obtain better instance representations. 
The framework also includes a method for aggregating cross-domain knowledge into models by 
open-source task enrichment.
Additionally, we construct a brand new dataset: the TinyRel-CM dataset, a few-shot relation 
classification dataset in health domain with purposely small training data
and challenging relation classes. Experimental results demonstrate that our framework brings
performance gains for most underlying classification models,
outperforms the state-of-the-art results given small training data, 
and achieves competitive results with sufficiently large training data.
\end{abstract}


\maketitle

\section{Introduction}
Relation classification (RC) is an indispensable problem in information extraction and knowledge discovery. Given a sentence (e.g., \emph{Washington is the capital of the United States}) containing two target entities (e.g., \emph{Washington} and \emph{the United States}), an RC model aims to distinguish the semantic relation between the entities mentioned by the sentence (e.g., \emph{capital of}) among all candidate relation classes.

Conventional relation classification has been extensively investigated. Recent approaches \cite{zeng-etal-2014-relation,RNNRC,vu-etal-2016-combining} train models with large amount of annotated data and achieve satisfactory results.
However, it is often costly to obtain necessary amount of labeled data by human-annotation,
without which a decrease in the performance of the RC models is inevitable.
In order to obtain sufficient labeled data,
distant supervised  methods \cite{NYTdataset} have been adopted to enlarge annotation quantity by utilizing existing knowledge bases to perform auto-labeling on massive raw corpus. However, long-tail problems \cite{xiong-etal-2018-one,han-etal-2018-fewrel,ye-ling-2019-multi} and noise issues occur.

\begin{table}[t]
	\centering
	\small
	\caption{\label{FewShotRCExample}
		An example of 5-way 5-shot relation classification scenario from the FewRel validation set. $Entity~1$ marks the head entity, and $entity~2$ marks the tail entity. The query instance is of Class 1: \emph{mother}. The support instances of Classes 2-5 are omitted.
	}
	\begin{tabular}{p{8cm}}
		\hline
		\textbf{Support Set} \\ \hline
		\textbf{Class 1} \emph{~mother}: \\
		\qquad \textbf{Instance\;1}~\lbrack Emmy Acht\'e\rbrack$_{entity~1}$ was the mother of the internationally famouse opera singers \lbrack Aino Ackt\'e\rbrack$_{entity~2}$ and Irma Tervani. \\
		\qquad \textbf{Instance\;2}~He was deposed in 922, and \lbrack Eadgifu\rbrack$_{entity~1}$ sent their son, \lbrack Louis\rbrack$_{entity~2}$ to safety in England. \\
		\qquad \textbf{Instance\;3}~Jinnah and his wife \lbrack Rattanbai Petit\rbrack$_{entity~1}$ had separated soon after their daughter, \lbrack Dina Wadia\rbrack$_{entity~2}$ was born. \\
		\qquad \textbf{Instance\;4}~Ariston had three other children by \lbrack Perictione\rbrack$_{entity~1}$: Glaucon, \lbrack Adeimantus\rbrack$_{entity~2}$, and Potone.\\
		\qquad \textbf{Instance\;5}~She married (and murdered) \lbrack Polyctor\rbrack$_{entity~2}$, son of Aegyptus and \lbrack Caliadne\rbrack$_{entity~1}$. Apollodorus.\\
		\textbf{Class 2} \emph{~part\_of}: ... \\
		\textbf{Class 3} \emph{~military\_rank}: ... \\
		\textbf{Class 4} \emph{~follows}: ... \\
		\textbf{Class 5} \emph{~crosses}: ... \\ \hline
		\textbf{Query Instance} \\ \hline
		Dylan and \lbrack Caitlin\rbrack$_{entity1}$ brought up their three children, \lbrack Aeronwy\rbrack$_{entity2}$, Llewellyn and Colm. \\
		\hline
	\end{tabular}
\end{table}

Few-shot relation classification is a particular RC task under minimum annotated data of 
{\em concerned relation classes}, i.e., a model is required to classify an incoming query 
instance given only few support instances (e.g., 1 or 5) during testing. 
An example is given in Table~\ref{FewShotRCExample}.
It is worth noting that, in previous few-shot RC tasks, although models see only few support instances during testing, they are trained with a large amount of data (e.g., tens of thousands of instances in total), labeled with relation classes different from but in the same domain of the {\em concerned
relation classes}.
This situation arises when instances of several relation classes are hard to find while 
those of others are abundant.
But in practice, it is often hard to get so many annotated training data as well. 
The difficulty 
lies in 2 aspects:
\begin{enumerate}
	\item Expertise required. Labeling is hard when it comes to professional fields such as bio-medical and risk management instead of general corpus. The relation classes in professional fields tend to be confusing and expertise is required.
	Thus the cost of labeling increases.
	\item Useless samples. Overwhelming majority (above 90\% in some of our experiments) 
of instances are of relation class \emph{NA} (no relation). 
This situation is particularly true in domains such as bio-medical and risk management, 
where a large proportion of instances are 
ordinary samples
but only the samples with health issues 
or risks are of our concern. This makes labeling very unproductive. 
\end{enumerate}
In this paper, we highlight the situation where in-domain training data is hard to obtain as is illustrated above, and restrict the size of training data in few-shot relation classification.

\begin{figure}[tbh!]
	\centering
	\includegraphics[width=8.5cm]{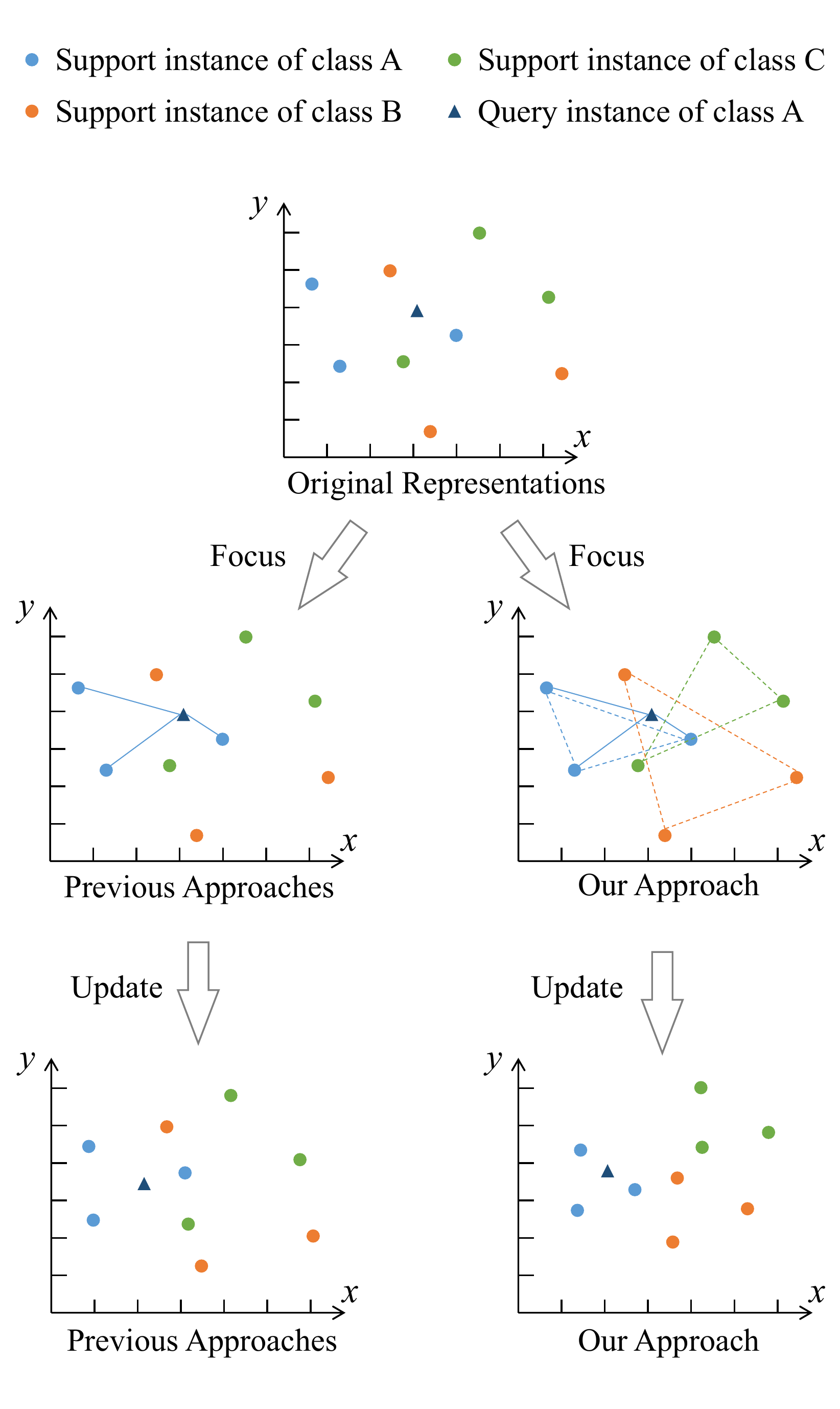}
	\caption{The general idea of model's updating process of one iteration. Previous approaches update the representations merely by the classification results on query instances. Our approach updates the representations by classification results on both query instances and support instances.
	Solid lines indicate the intention of making representations of the query instance and each support instance of the same class closer to each other.
	Dotted lines indicate the intention of making representations of support instances within each class closer.
	}
	\label{fig:support}
\end{figure}

Meta-learning is a popular method for few-shot learning circumstances and is broadly studied in computer vision (CV) \cite{LakeHuman,Santoro2016,proto}.
Instead of training a neural network to learn a specific task, in meta-learning, the model is trained with a variety of similar but different tasks to gain the ability of quick adaptation to new tasks without meeting a ton of data.
One typical framework of meta-learning is to train an additional meta-learner %
which sets and adjusts the update rules for the conventional learner \cite{Andry2016,Finn2017,HN}. Another framework is based on metric learning and is aimed at
learning the distance distribution among the relation classes \cite{Koch2015,Vinyals2016,proto}.
In recent years, meta-learning has been shown to help in few-shot NLP tasks,
including few-shot relation classification, which is the focus of this paper.
Han et al. \shortcite{han-etal-2018-fewrel} constructed the FewRel dataset, a few-shot relation classification dataset and applied distinct meta-learning frameworks intended for CV tasks on the FewRel dataset. Ye and Ling \shortcite{ye-ling-2019-multi}, Gao et al. \shortcite{hatt} and Gao et al. \shortcite{gao-etal-2019-fewrel} improved the models specifically for relation classification task
and achieved better performance.

While meta-learning frameworks outperform conventional methods in few-shot RC, 
they are not applicable to situations where only small amount of annotated training data is available, which does happen in practice.
In previous work, although only few support instances are needed during testing, the training set still must be sufficiently large (e.g., the FewRel dataset \cite{han-etal-2018-fewrel} contains 700 instances per class). Performance drops significantly when the training data size is restricted
(e.g., to tens of instances per relation). Strong baselines such as MLMAN \cite{ye-ling-2019-multi} and Bert-Pair \cite{gao-etal-2019-fewrel} achieves satisfactory performance (about 80\% accuracy on 5-way 1-shot tasks) with full FewRel training data, but accuracy decreases by 20\% if 10 relation classes with only 10 instances per class are given as training data (see Section \ref{results}).
Besides, previous meta-learning methods concern much about computations on query instances but lose sight of knowledge within the support instances.
Typical methods regard the features of support instances as \emph{standards} to classify query instances during training iterations.
We hold the view that improving the quality of the \emph{standards} itself
by extracting knowledge within support instances is important but
overlooked by previous work.

To 
enhance models and make them better cater to small training data, we propose
\emph{Meta-learning using Intra-support and Cross-domain Knowledge} (MICK) framework.
First, we
utilize cross-domain knowledge by enriching the training tasks with cross-domain relation classification dataset
(e.g., adding relation class \emph{mother} to a medical relation classification training set,
and forming a 3-way training task with relation classes \emph{Disease-cause-Disease}, \emph{Disease-have-Symptom}, and \emph{mother})
(\secref{sec:data}).
The only requirement on the cross-domain dataset is to share the same language with the original dataset, thus it is easy to find.
Although differences exist in distributions of data from distinct domains, basic knowledge such as language principles and grammar are shared, and thus compensate for the insufficient learning of basic knowledge due to lack of training data.
Moreover,
even adding only a few cross-domain relation classes brings about a huge increase in the amount of feasible tasks. Task enrichment makes the model more reliable by forcing it to solve extensive and diverse training tasks.
Second, inspired by Chen et al. \shortcite{chen-2019-image}, we exploit underlying knowledge within support instances using a support classifier scheduled by a fast-slow learner (\secref{sec:cls}).
Instead of updating the instance representations merely by classifying the query instances according to the support instances, 
we also use the support classifier to classify the support instances and update the model so that representations of support instances of the same class are closer to each other (see \figref{fig:support}). 
This makes the instance representations within the same class more compact, thus the classification of query instances becomes more accurate.

Additionally, we propose our own dataset, TinyRel-CM dataset,
a Chinese few-shot relation classification dataset in health domain. 
Different from previous few-shot RC dataset which contains abundant training data, we purposely limit the size of training data of our dataset.
The TinyRel-CM dataset contains 27 relation classes with 50 instances per class, 
which sets up the challenge of few-shot relation classification under \emph{small} training data. 
This dataset
is also challenging because the relation classes in the test set are all very similar to each
other. 
Experiments are conducted on both our proposed TinyRel-CM dataset and the FewRel dataset \cite{han-etal-2018-fewrel}. Experimental results show the strengths of our proposed framework.


In summary, our contributions include:
 (1) We propose a meta-learning framework that achieves state-of-the-art performance with the limitation of small training data and competitive results with sufficient training data (\secref{results}).
(2) We utilize a support classifier to extract intra-support knowledge and
obtain more reasonable instance representations and demonstrate the improvement (\secref{results}).
(3) We propose a task enrichment method to utilize cross-domain implicit knowledge, which is extremely useful under small training data (\secref{results}).
(4) We propose TinyRel-CM dataset,
the {\bf second} and a challenging dataset for few-shot relation classification task with small training data (\secref{dataset} and \secref{results}).


\section{Problem Formulation}
We add limitation on the size of training data compared with previous few-shot relation classification tasks.
Just like in conventional few-shot relation classification, there is a training set
$D_{\rm{train}}$ and a test set $D_{\rm{test}}$.
Each instance in both sets can be represented as a triple $(s,e,r)$,
where $s$ is a sentence of length $T$, $e=(e_1, e_2)$ is the head and tail entities and
$r$ is the semantic relation between $e_1$ and $e_2$ conveyed by $s$.
$r \in R$, where $R=\{r_1,...,r_N\}$ is the set of all candidate relation classes.
$D_{\rm{train}}$ and $D_{\rm{test}}$ have disjoint relation sets, i.e.,
if a relation $r$ appears in a triple of the training set, it must not appear
in any triples of the test set and vice versa.
$D_{\rm{test}}$ is further split into a support set $D_{\rm{test\text{-}s}}$
and a query set $D_{\rm{test\text{-}q}}$. The problem is to predict
the classes of instances in $D_{\rm{test\text{-}q}}$ given
$D_{\rm{test\text{-}s}}$ and $D_{\rm{train}}$. While no restrictions are lied on how to use $D_{\rm{train}}$, it is conventionally splited into a support set and a query set to train models.
In a $N$-way $K$-shot 
scenario,
$D_{\rm{test\text{-}s}}$ contains $N$ relation classes and $K$ instances
for each class. Both $N$ and $K$ are supposed to be small (e.g., 5-way 1-shot, 10-way 5-shot).
Particularly, we limit the size of training data (i.e.,  $D_{\rm{train}}$ is also small).
The difficulty of 
the task lies in not only the small size of
$D_{\rm{test\text{-}s}}$ (totally $N\times K$ instances) but also the small training data size.



\section{Methodology}
\begin{figure*}[ht]
    \centering
    \includegraphics[width=17cm]{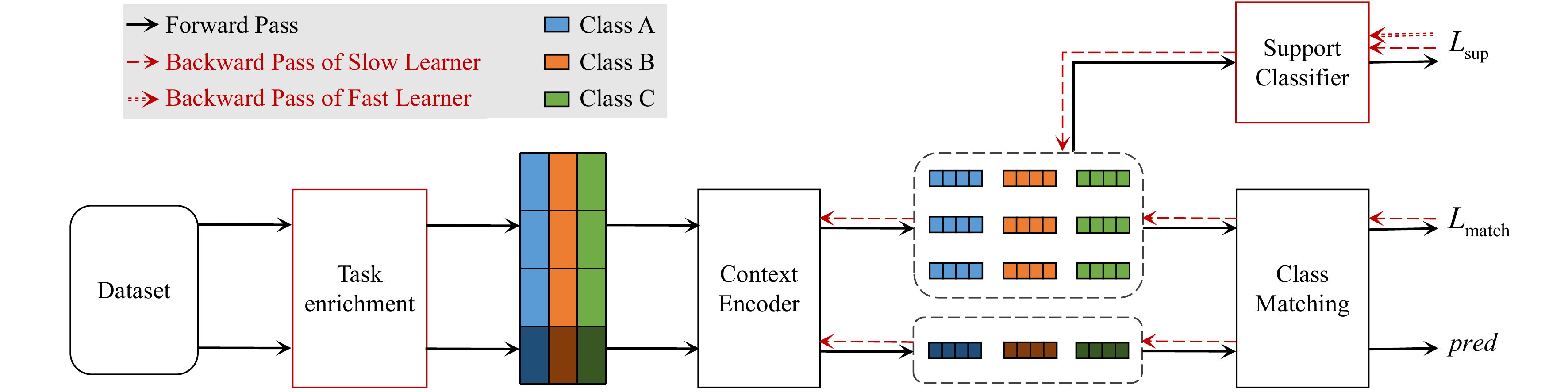}
    \caption{The structure and learning process of the MICK framework (under a 3-way 3-shot example). Modules with black border are typical meta-learning components. Modules with red border are our improvement. Cells of different colors represent instances from different classes. Light colors represent support instances. Dark colors represent query instances.
    }
    \label{fig:model}
\end{figure*}

The proposed MICK framework contains a task enrichment method to
aggregate cross-domain knowledge, and a support classifier scheduled by
a fast-slow learner strategy to extract intra-support knowledge (illustrated in Figure \ref{fig:model}).
Next we present the structure and the learning process of MICK.

\label{sec:Methodology}

\subsection{Overview of MICK}
\label{sec:structure}
As is shown in Figure \ref{fig:model},
the structure of our framework
consists of four main parts: {\em task enrichment}, {\em context encoder}, {\em class matching}
and {\em support classifier}.

During training with task enrichment, for each episode, we randomly select a 
$N$-way $K$-shot task composed of $N \times K$ support instances and several query instances extracted from both the original training set $D_{\rm{train}}$ and supplementary cross-domain dataset.
Task enrichment enables the context encoder to learn cross-domain underlying knowledge.
The instances are first fed into the context encoder, which generates representation vectors for each instance.
Then we forward the encoded support and query vectors to class matching. Class matching aims to classify the query instances according to the representations of support instances.
Additionally, the support vectors are fed into a support classifier, which is a $N$-way linear classifier.
The support classifier aims to extract knowledge within support instances to facilitate the context encoder.
The back propagation process is scheduled by a fast-slow learner. Fast-slow learner scheme is motivated by meta-learner based meta learning methods \cite{Andry2016,Finn2017,HN} where the traditional learner and meta-learner learn with different speed. Task-specific parameters in the support classifier update every episode with a fast learner. Task-agnostic parameters in the context encoder update every $\epsilon$ episodes with a slow learner.
With the fast-slow learner, we obtain a support classifier which can quickly adapt to new classification tasks and a global context encoder that fits all tasks.

During testing, we only use the context encoder and class matching method to make predictions on $D_{\rm{test}}$.

\subsection{Preliminaries}
Modules in black boxes in Figure \ref{fig:model} compose an ordinary meta-learning framework.

For an instance $(s,e,r)$, sentence $s=\{c_0,..., c_i, ... c_{T-1}\}$ is padded to a predefined 
length $T$, where $c_i$ represents the one-hot vector of the $i^{th}$ word.
In context encoder, each $c_i$ is mapped to an embedding $\mathbf{x}_i \in \mathbb{R}^{d_c + 2 d_p}$, where $d_c$ is the word embedding\footnote{Character embedding for Chinese corpus.} size and $d_p$ is the position embedding size.
The representation matrix of sentence $s$ is the concatenation of $\mathbf{x}_i$: $\mathbf{X}=[\mathbf{x}_0,..., \mathbf{x}_i, ... ,\mathbf{x}_{T-1}]$. $\mathbf{X} \in \mathbb{R}^{T\times (d_c + 2  d_p)}$.

$\mathbf{X}$ is further fed into a sentence encoder (e.g., CNN or LSTM) to extract semantics of sentence $s$ and get representation vector $\mathbf{E} \in \mathbb{R}^{d_h}$, where $d_h$ is the dimension of hidden states.
Thus, an instance can be represented as a pair $(\mathbf{E},r)$, where $\mathbf{E}$ is the representation vector and $r$ is the relation class.


Given an encoded query instance $Q=(\mathbf{E}^q, r^q)$ where $r^q$ is to be predicted, class matching aims to match $r^q$ with some relation class $r_i \in R$. Conventionally, a function $\mathcal{F}$ is adopted to measure the distance between $\mathbf{E}^q$ and $\mathbf{E}^{i}$, where $\mathbf{E}^{i}$ is the representation vector of relation $r_i$ and is calculated with all the representation vectors of support instances belonging to class $r_i$.
$\mathcal{F}$ can be a function either without parameters (e.g., Euclidean distance function) or with parameters (e.g., a linear classifier).
Relation class $r_i$ is chosen as the prediction class if $\mathbf{E}^i$ has the closest distance to $\mathbf{E}^q$.

\subsection{Task Enrichment}
\label{sec:data}
In order to expand the size of training data and enrich training tasks, we utilize cross-domain relation classification datasets in the same language. These datasets are obtained from released data of other works or online resources. This task enrichment step is necessary under the circumstance of a tiny training 
dataset.
With cross-domain datasets, training tasks are randomly extracted from (1) original data, (2) cross-domain data and original data, (3) original data, sequentially.
This three-phase training scheme simulates the process of students learning from the simple to the deep, and reviewing before exams.

\subsection{Support classifier and fast-slow learner}
\label{sec:cls}
To further explore knowledge within support instances, we introduce a support classifier. The classifier receives representation vector $\mathbf{E}$ of each support instance as input and outputs the the probability of the support instance belonging to each relation class. 
The output $\mathbf{O}$ equals to
\begin{eqnarray}
\mathbf{O} &=& \frac{exp(\mathbf{M}[i])}{\sum_{j=0}^{N-1} exp(\mathbf{M}[j])}, \\
\mathbf{M} &=& \mathbf{WE}+ \mathbf{b},
\end{eqnarray}
where $\mathbf{W}$ and $\mathbf{b}$ are parameters to be trained, and $\mathbf{M}[i]$ represents the $i^{th}$ element of $\mathbf{M}$.



\begin{algorithm}[th]
\small
\caption{Meta-learning with Support Classifier}
\label{alg:metal}
\leftline{\textbf{Require:}
distribution over relation classes in training set $p(\mathcal{R})$, context}
\leftline{encoder $\mathcal{E}_{\theta_{slow}}$,
class matching function $\mathcal{F}$,
support classifier $\mathcal{G}_{\theta_{fast}}$,}
\leftline{fast learner learning rate $\alpha$,
slow learner learning rate $\beta$,
step size $\epsilon$,}
\leftline{\#classes per task $N$}
\begin{algorithmic}[1]
\STATE Randomly initialize task-specific parameters $\theta_{fast}$
\STATE Randomly initialize task-agnostic parameters $\theta_{slow}$
\WHILE {not done}
\STATE Initialize slow loss $\mathcal{L}_{slow}=0$
\FOR {$j=1$ to $\epsilon$}
\STATE Sample $N$ classes $\mathcal{R}_i \thicksim p(\mathcal{R})$ from training set $D_{\rm{train}}$
\label{sampler}
\STATE Sample instances $\mathcal{H}=(s^{(i)}, e^{(i)}, r^{(i)})$ from $\mathcal{R}_i$
\label{samplei}
\STATE Compute $\mathcal{L}_{sup}(\mathcal{E}_{\theta_{slow}}, \mathcal{G}_{\theta_{fast}})$ using $\mathcal{H}$
\label{Lsup}
\STATE Compute $\mathcal{L}_{match}(\mathcal{E}_{\theta_{slow}}, \mathcal{F})$ using $\mathcal{H}$
\STATE Fast loss $\mathcal{L}_{fast}=\mathcal{L}_{sup}$
\label{fastloss}
\STATE $\mathcal{L}_{slow} = \mathcal{L}_{slow} + \mathcal{L}_{sup} + \mathcal{L}_{match}$
\label{slowloss}
\STATE $\theta_{fast} = \theta_{fast} - \alpha \bigtriangledown_{\theta_{fast}} \mathcal{L}_{fast}$
\label{fast}
\ENDFOR
\STATE $\theta_{slow} = \theta_{slow} - \beta \bigtriangledown_{\theta_{slow}} \mathcal{L}_{slow}$
\label{slow}
\ENDWHILE
\STATE Use $\mathcal{E}_{\theta_{slow}}$ and $\mathcal{F}$ to perform classification of test set $D_{\rm{test}}$.
\label{test}
\end{algorithmic}
\end{algorithm}

The learning process with the support classifier is scheduled by a fast-slow learner (see Algorithm \ref{alg:metal}).
The training process contains multiple episodes. For each episode, a task is randomly generated from the training data (line \ref{sampler}, \ref{samplei}).
During training, we train two learners: a fast learner with learning rate $\alpha$ and a slow learner with learning rate $\beta$.

The fast learner learns parameters of the support classifier which are task-specific, and updates after each episode (line \ref{fast}). The objective function of the fast learner is cross entropy loss (line \ref{Lsup}, \ref{fastloss}):
\begin{equation}
\begin{aligned}
    \mathcal{L}_{fast} = \mathcal{L}_{sup}= - \sum_{i=0}^{N-1} \mathbf{r}[i] log(\mathbf{O}[i]), \\
\end{aligned}
\end{equation}
where $N$ is the number of classes, $\mathbf{r}[i]$ is the $i^{th}$ element of ground truth one-hot vector $\mathbf{r}$, and $\mathbf{O}[i]$ is the $i^{th}$ element of the output of the support classifier $\mathbf{O}$.

The slow learner learns parameters of the context encoder (line \ref{slowloss}) with objective function:
\begin{equation}
\mathcal{L}_{slow} = \mathcal{L}_{sup} + \mathcal{L}_{match},
\end{equation}
where $\mathcal{L}_{match}$ is the objective function inherited from the core model which provides the context encoder and the class matching function. $\mathcal{L}_{slow}$ accumulates during every $\epsilon$ episodes and then back propagates (line \ref{slowloss}, \ref{slow}).

\section{Experiments}
\label{exp}
In Section \ref{exp}, we explain the experiment process in detail, including datasets, experimental setup, implementation details and evaluation results.

\subsection{Dataset}
\label{dataset}
Experiments are done on two datasets: FewRel dataset \cite{han-etal-2018-fewrel} and our proposed TinyRel-CM dataset.
~\\
~\\
\textbf{FewRel Dataset} FewRel dataset \cite{han-etal-2018-fewrel} is a few-shot relation classification dataset constructed through distant supervision and human annotation. It consists of 100 relation classes with 700 instances per class. The relation classes are split into subsets of size 64, 16 and 20 for training, validation and testing, respectively.
The average length of a sentence in FewRel dataset is 24.99, and there are 124,577 unique tokens in total. At the time of writing, FewRel is the only few-shot relation classification dataset available.
~\\
~\\
\textbf{TinyRel-CM Dataset}\footnote{Code and dataset released in \url{https://github.com/XiaoqingGeng/MICK}.} TinyRel-CM dataset is our proposed Chinese few-shot relation classification dataset in health domain with small training data. The TinyRel-CM dataset is constructed through the following steps: (1) Crawl data from Chinese health-related websites\footnote{\url{www.9939.com}, \url{www.39.net}, and \url{www.xywy.com}} to form a large corpus and an entity dictionary. (2) Automatically align entities in the corpus with the entity dictionary, forming a large candidate-sentence set. (3) 5 Chinese medical students manually filter out the unqualified candidate sentences and tag qualified ones with corresponding class labels to form an instance. An instance is added to the dataset only if 3 or more annotators make consistent decisions. This process costs 4 days.

The TinyRel-CM Dataset consists of 27 relation classes with 50 instances per class. The 27 relation classes cover binary relations among 4 entity types, and are grouped into 6 categories according to the entity types, forming 6 tasks with one group being the test set and other 5 groups serving as training set (see Table \ref{Egroup}). Grouping makes TinyRel-CM dataset more challenging because all candidate relation classes during testing are highly similar. An example instance in the TinyRel-CM dataset is shown in Tabel \ref{FRMexample}. The average length of a sentence in TinyRel-CM dataset is 67.31 characters, and there are 2,197 unique characters in total. Comparison between TinyRel-CM dataset and FewRel dataset is shown in Table \ref{Datasetcompare}.

\begin{table}[ht]
\centering
\small
\caption{Entity groups in TinyRel-CM dataset. D,S,F, and U stand for Disease, Symptom, Food, and nUtrient, respectively.}
\label{Egroup}
\begin{tabu}{|c|[0.5pt]l|}
\hline
\textbf{Group} & \textbf{Classes} \\ \tabucline[0.5pt]{-}
D-D & complication, cause, is, include, NA \\ \hline
D-S & have, NA\\ \hline
D-F & positive, negative, forbid, prevent, cause, NA\\ \hline
D-U & positive, negative, prevent, lack, cause, NA\\ \hline
F-U & contain, NA\\ \hline
S-F & forbid, cause, positive, negative, prevent, NA\\ \hline
\end{tabu}
\end{table}

\begin{table}[ht]
\centering
\small
\caption{An example instance in the TinyRel-CM dataset.}
\label{FRMexample}
\begin{tabular}{|l|p{170pt}|}
\hline
\textbf{Group} & D-D \\ \hline 
\textbf{Class} & complication \\ \hline
\textbf{Explanation} & Entity 2 is a complication of entity 1. \\ \hline
\multirow{2}{*}{\textbf{Example}} & [子宫肌瘤]$_{entity1}$出现了[慢性盆腔炎]$_{entity2}$并发症，导致月经量过多。 \\
& \textcolor[rgb]{0.45,0.45,0.45}{Once [Hysteromyoma]$_{entity1}$ is complicated with [chronic pelvic inflammation]$_{entity2}$, menstruation increases.}  \\ \hline
\end{tabular}
\end{table}

\begin{table}[ht]
\centering
\small
\caption{Comparison of TinyRel-CM dataset to FewRel dataset.}
\label{Datasetcompare}
\begin{tabular}{|l|r|r|r|}
\hline
\textbf{Dataset} & \textbf{\#cls.} & \textbf{\#inst./cls.} & \textbf{\#inst.} \\ \hline
FewRel & 100 & 700 & 70,000 \\ \hline
TinyRel-CM & 27 & 50 & 1,350 \\ \hline
\end{tabular}
\end{table}

\subsection{Experimental Setup}

We first conduct experiments with small training data.
On the TinyRel-CM dataset, for each group of relation classes, we adopt $N$-way 5-shot, $N$-way 10-shot and $N$-way 15-shot test configurations, where $N$ is the number of classes within the group. During training episodes, we conduct 5-way 15-shot training tasks. Thus totally 6 experiments are done on TinyRel-CM dataset.
On the FewRel dataset, we modify the training set by shrinking the number of relation classes and instances per class to various extent. This aims to show not only the effect of our framework under small training data but also the performance trends of models with the change of data size.
For each shrunken training set, we conduct different training task settings
(shown in Table \ref{trainingsetting}) and test with 4 configurations:
5-way 1-shot, 5-way 5-shot, 10-way 1-shot and 10-way 5-shot. 

\begin{table}[ht]
\centering
\small
\caption{Training task settings over shrunken training set.}
\label{trainingsetting}
\begin{tabular}{|c|c|c|l|}
\hline
\textbf{\% of full training set} & \textbf{\#cls.} & \textbf{\#inst./cls.} & \textbf{Training task} \\ \hline
7.00 & 30 & 100 & ~5\text{-}way 15\text{-}shot  \\ \hline
2.23 & 20 & 50 & ~5\text{-}way 15\text{-}shot \\ \hline
1.00 & 15 & 30 & ~5\text{-}way 10\text{-}shot \\ \hline
0.22 & 10 & 10 & ~5\text{-}way ~~~5\text{-}shot \\ \hline
\end{tabular}
\end{table}

Second, we experiment with sufficient training data.
On the FewRel dataset, following Han et al. \shortcite{han-etal-2018-fewrel} and Ye and Ling \shortcite{ye-ling-2019-multi}, we train the model with 20-way 10-shot training tasks and test with 4 configurations: 5-way 1-shot, 5-way 5-shot, 10-way 1-shot and 10-way 5-shot.

For the ablation tests, in addition to applying the whole MICK framework, we also apply the two proposed methods, support classifier and task enrichment, individually on baseline models.

For all experiments, we randomly pick 2000 tasks and calculate the average accuracy in testing.


\subsection{Implementation Details}

During implementation, we apply our framework and data augmentation method to the following baselines:
GNN \cite{gnn},
SNAIL \cite{snail},
prototypical networks \cite{proto},
proto-HATT \cite{hatt},
MLMAN \cite{ye-ling-2019-multi}, and
Bert-Pair \cite{gao-etal-2019-fewrel}.

Codes for GNN, SNAIL and Bert-Pair are provided by Gao et al. \shortcite{gao-etal-2019-fewrel}. Prototypical network uses our own implementation. Codes for proto-HATT and MLMAN are provided in the original paper.

Due to the particularity of the Bert-Pair model, the support classifier applied on Bert-Pair receives support instance pairs as input and computes the probability of the pairs belonging to the same class, different from other baselines. Although the distinct implementation of support classifier, we keep our intention to extract knowledge within support instances.

GNN, SNAIL, and proto-HATT require the number of classes while training and testing to be equal. So a model is trained with $N$-way tasks to perform $N$-way test tasks. In SNAIL and proto-HATT, the number of instances per class while training and testing need to be equal. So a model is trained with $K$-shot tasks to perform $K$-shot test tasks.

We keep the original hyper parameters for each baseline, and set the learning rate of the fast learner $0.1$. The cross-domain data for TinyRel-CM dataset are from Chinese Literature NER RE dataset \cite{dnerre} (13,297 instances covering 10 classes in general corpus including \emph{part\_whole}, \emph{near}, etc.) and Chinese Information Extraction dataset \cite{augdata} (1,100 instances covering 12 classes between persons including \emph{parent\_of}, \emph{friend\_of}, etc.). Cross-domain data for FewRel dataset is from NYT-10 dataset \cite{NYTdataset} which contains 143,391 instances over 57 classes in general courpus including \emph{contain}, \emph{nationality}, etc. (class \emph{NA} is removed during task enrichment).
The only requirement on the supplementary dataset is to share the common language with the original dataset.

\begin{figure*}[th]
	\centering
	\small
	\subfigure[5-way 1-shot]{
		\centering
		\includegraphics[width=\linewidth]{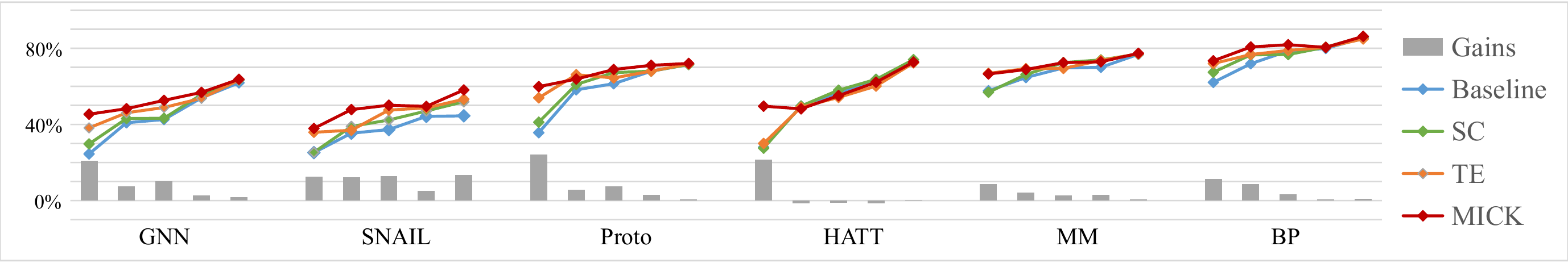}
	} \\
	\subfigure[5-way 5-shot]{
		\centering
		\includegraphics[width=\linewidth]{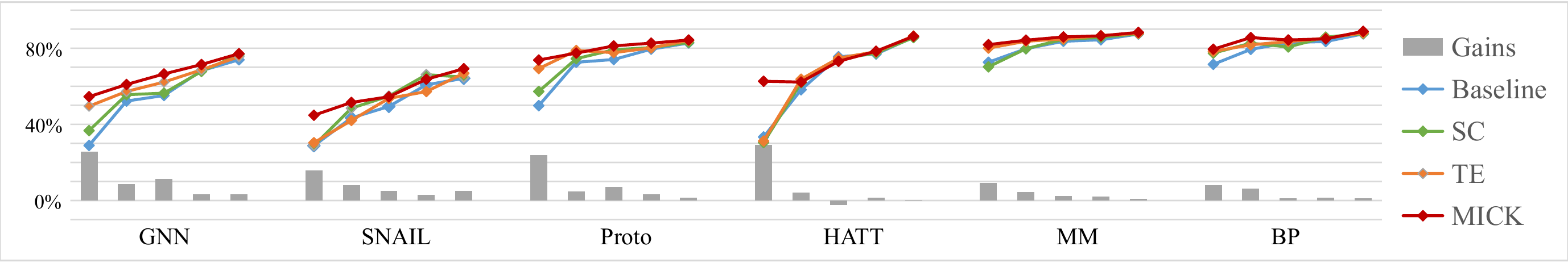}
	} \\%
	\subfigure[10-way 1-shot]{
		\centering
		\includegraphics[width=\linewidth]{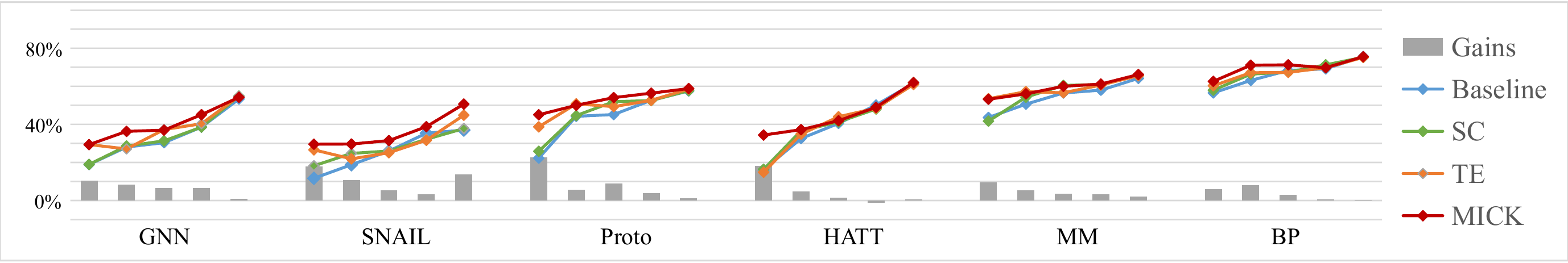}
	} \\
	\subfigure[10-way 5-shot]{
		\centering
		\includegraphics[width=\linewidth]{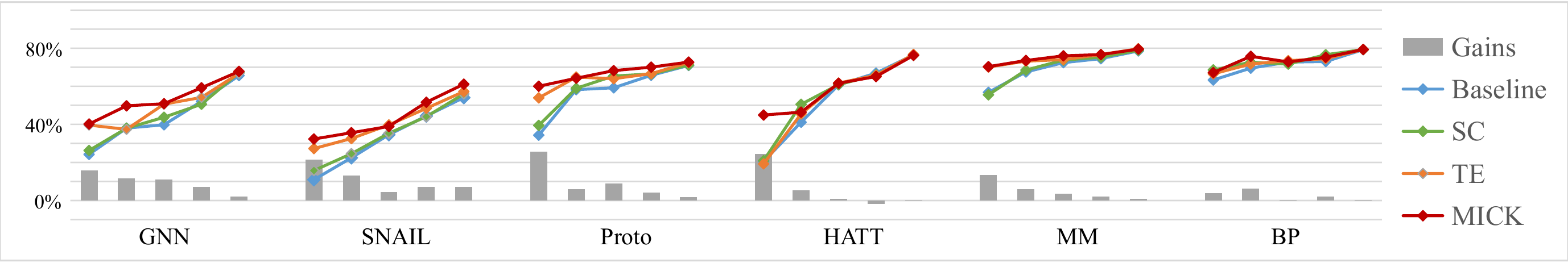}
	}%
	
	\centering
	\caption{Classification accuracy on FewRel validation set 
		under $N$-way $K$-shot configurations. Gains is the difference between MICK accuracy and baseline accuracy.
		For each group, from the left to the right, training data is shrunken to 0.22\%, 1.00\%, 2.23\%, 7.00\%, and 100.00\% of full training set size, respectively. For each shrunken training set, we apply baseline models and models with SC (Support Classifier individually), TE (Task Enrichment individually) and the whole MICK framework. Proto, HATT, MM, and BP stand for Prototypical Networks, Proto-HATT, MLMAN, and Bert-Pair, respectively.}
	\label{fig:analysis}
\end{figure*}

\subsection{Results and Analysis}
\label{results}
Here, we show the experimental results and analyze them from different aspects.

\subsubsection{With small training data}
To show the effectiveness of our MICK framework under small training data,
we apply it on each baseline model on
(1) FewRel dataset with training set shrunken to different extents, and (2) our proposed TinyRel-CM dataset.

\begin{table*}[th]
\centering
\small
\caption{Classification accuracy (\%) on TinyRel-CM dataset under
	$N$-way $K$-shot configuration.
	D, S, F, and U stand for Disease, Symptom, Food, and nUtrient, respectively.
	Proto, HATT, MM, and BP stand for Prototypical Networks, Proto-HATT, MLMAN, and Bert-Pair, respectively.
	Gray numbers indicate the accuracy is lower than baseline.}
\label{FRMresult}
\begin{tabular}{|c|c|cccccc|cccccc|}
\hline
\textbf{Method} & &\multicolumn{6}{c|}{\emph{Baseline}} &\multicolumn{6}{c|}{\emph{+SupportClassifier}} \\ \hline
&\textbf{Group} & D-D & D-S & D-F & D-U & F-U & S-F& D-D & D-S & D-F & D-U & F-U & S-F\\
&\textbf{(N)} & (5) & (2) & (6) & (6)& (2) & (6)& (5) & (2) & (6) & (6)& (2) & (6)\\ \hline
\multirow{3}{*}{GNN}   & (K=~5~)   & 24.04&53.62 &34.15 &38.90 & 54.60& 32.45 &25.71&57.12&33.11&40.23&\textcolor[rgb]{0.45,0.45,0.45}{53.20}&34.51 \\
 & (K=10) & 26.05& 53.73&37.54 &44.02 &\textbf{55.53} &35.45 &\textbf{27.96}&57.35&37.44&43.87&\textcolor[rgb]{0.45,0.45,0.45}{54.85}&\textbf{38.39} \\
 & (K=15) & 26.94& 54.30&38.36 &45.12 &55.45 &37.43 &\textbf{28.36}&61.25&39.18&47.57&\textcolor[rgb]{0.45,0.45,0.45}{55.10}&\textbf{40.37}\\ \hline
\multirow{3}{*}{SNAIL}
 & (K=~5~) &21.74 & 50.00 &17.78 &25.98 & 47.96& 25.38     &23.64&53.90&27.28&27.60&53.15&25.88\\
  & (K=10) & 19.49 & 53.57&27.55 &22.96 &51.63 &22.51      &23.98&56.65&27.76&30.10&54.60&26.10 \\
 & (K=15) & 21.94& 51.62&20.03 &18.21 &50.33 &21.38         &23.12&\textbf{57.40}&27.72&31.43&\textbf{56.35}&25.40\\ \hline
\multirow{3}{*}{Proto}      & (K=~5~)  & 30.03&63.27 &31.35 &33.95 & 54.99& 26.97 &28.99 &\textbf{69.56} &33.57 & 43.78&56.49 &36.81\\
 & (K=10) & 31.68& 67.22&34.09 &36.27 &56.78 &29.18 &32.35 &\textbf{72.54} &36.10 & 47.78& 58.13&39.04\\
 & (K=15) & 33.80& 70.29&34.57 &37.41 &57.03 &29.72 &34.82 &\textbf{72.79} &36.96 &\textbf{50.03} &\textbf{ 60.09} &40.09\\ \hline

\multirow{3}{*}{HATT}
 & (K=~5~) &29.67&59.27&40.13&40.19&65.20&41.67  
&36.39&71.45&40.83&\textbf{54.17}&70.09&43.33 \\
 & (K=10) &34.50&59.77&39.17&48.33&67.50&43.57  
&40.10&\textbf{71.24}&45.34&50.83&73.33&\textbf{49.17} \\
 & (K=15) &39.01&63.75&37.41&\textbf{56.67}&66.25&37.13  
&43.05&77.09&44.17&\textcolor[rgb]{0.45,0.45,0.45}{51.67}&\textcolor[rgb]{0.45,0.45,0.45}{63.75}&45.92  
\\
\hline

\multirow{3}{*}{MM}   & (K=~5~)     &31.22 &60.58 &44.97 &47.11 &61.04 &44.78 &33.48 &62.34 & \textcolor[rgb]{0.45,0.45,0.45}{44.93}&49.49 &62.40 &45.90\\
 & (K=10) & 36.66&64.96 &49.44 &52.49 &66.68 &50.57 &38.89 & 67.30& 50.09& 54.09&68.39 &50.77\\
 & (K=15) & 40.65&68.14 &52.44 &56.00 &70.32 &53.13 &43.05 & 69.41&53.82 &57.23 &72.02 & \textcolor[rgb]{0.45,0.45,0.45}{52.87}\\ \hline
\multirow{3}{*}{BP}  & (K=~5~)    & 23.61&53.48 &46.02 &42.03 & 54.17& 46.18&   24.60 &\textcolor[rgb]{0.45,0.45,0.45}{53.98}&50.49 &\textbf{43.28}&\textbf{63.33}&48.75 \\
 & (K=10) & 24.17& 56.95&49.52 &44.37 &59.72 &48.96 &26.69&\textcolor[rgb]{0.45,0.45,0.45}{55.05} &53.60 &\textbf{46.13}&\textbf{66.70} &51.63  \\
 & (K=15) & 25.80& 54.15&50.38 &45.16 &59.33 &48.78 &27.29 &56.65 &54.22&\textbf{47.81} &67.55 &52.48 \\ \hline

\textbf{Method} & & \multicolumn{6}{c|}{\emph{+TaskEnrich}} &\multicolumn{6}{c|}{\emph{+SupportClassifier\&TaskEnrich}} \\ \hline
& \textbf{Group} & D-D & D-S & D-F & D-U & F-U & S-F& D-D & D-S & D-F & D-U & F-U & S-F\\
& \textbf{(N)} & (5) & (2) & (6) & (6)& (2) & (6)& (5) & (2) & (6) & (6)& (2) & (6)\\ \hline
\multirow{3}{*}{GNN}
 & (K=~5~) & 24.59&\textcolor[rgb]{0.45,0.45,0.45}{51.20}&\textcolor[rgb]{0.45,0.45,0.45}{33.28}&40.76&54.62&35.36 &\textbf{26.02}&\textbf{66.22}&\textbf{37.48}&\textbf{44.47}&\textbf{55.02}&\textbf{37.13} \\
 & (K=10) &26.85&56.05&37.56&44.90&\textcolor[rgb]{0.45,0.45,0.45}{54.12}&37.69 &27.66&\textbf{58.75}&\textbf{37.88}&\textbf{45.15}&\textcolor[rgb]{0.45,0.45,0.45}{53.42}&37.85 \\
 & (K=15) &27.71&58.55&40.42&49.43&56.35&40.13 &27.40&\textbf{70.10}&\textbf{43.25}&\textbf{51.40}&\textbf{56.45}&40.03 \\ \hline
\multirow{3}{*}{SNAIL}
 & (K=~5~) &22.43&53.95&25.17&\textbf{29.05}&51.30&27.27&\textbf{24.64}&\textbf{58.55}&\textbf{28.57}&27.08&\textbf{53.30}&\textbf{27.88} \\
 & (K=10) &22.79&\textcolor[rgb]{0.45,0.45,0.45}{51.98}&\textbf{30.67}&\textcolor[rgb]{0.45,0.45,0.45}{22.13}&53.15&23.03         &\textbf{27.38}&\textbf{57.12}&\textbf{27.85}&\textbf{33.50}&\textbf{55.60}&\textbf{30.52} \\
 & (K=15) &\textcolor[rgb]{0.45,0.45,0.45}{21.54}&53.10&20.77&\textbf{20.35}&52.00&23.30     &\textbf{23.84}& 56.70&\textbf{30.63}&\textbf{35.75}&54.30&\textbf{30.43} \\ \hline

\multirow{3}{*}{Proto}
 & (K=~5~) &31.40 &69.22 &32.13& 34.22&55.34&30.34&\textbf{31.50} &68.79 &\textbf{34.36} & \textbf{44.76} &\textbf{57.43}&\textbf{40.36} \\ 
 & (K=10) &34.80 &72.34 &34.51&36.58&57.29&33.81&\textbf{35.45} &71.37 &\textbf{37.11} & \textbf{48.28}&\textbf{58.71}&\textbf{43.55} \\
 & (K=15) &36.97 &71.62 &35.76 &37.79 &57.74&34.49&\textbf{38.28} &72.62 &\textbf{37.90} &49.58 &60.00&\textbf{44.97} \\ \hline

\multirow{3}{*}{HATT}
 & (K=~5~) &33.70&67.50&41.67&43.33&65.62&\textcolor[rgb]{0.45,0.45,0.45}{38.33} &\textbf{38.21}&\textbf{75.00}&\textbf{46.67}&49.17&\textbf{73.75}&\textbf{44.17} \\
 & (K=10) &38.98&68.75&40.13&48.75&\textcolor[rgb]{0.45,0.45,0.45}{60.92}&46.67  
&\textbf{44.45}&70.23&\textbf{49.37}&\textbf{51.67}&\textbf{75.00}&44.47 \\
 & (K=15) &40.39&72.31&43.84&\textcolor[rgb]{0.45,0.45,0.45}{55.83}&70.51&43.33  
&\textbf{49.61}&\textbf{80.41}&\textbf{44.33}&\textcolor[rgb]{0.45,0.45,0.45}{55.81}&\textbf{72.58}&\textbf{49.17} \\ \hline

\multirow{3}{*}{MM}
 & (K=~5~) &36.05 & \textbf{64.98}& 45.24&50.03 &61.34 & 46.82&\textbf{38.17} &64.68 & \textbf{45.32}& \textbf{51.30}& \textbf{64.24}& \textbf{48.34} \\
 & (K=10) &42.04 & \textbf{70.02}& \textbf{50.83} &54.85 &67.57 &52.47&\textbf{44.89} &69.93 & 50.75& \textbf{56.60}& \textbf{70.79} &\textbf{53.69} \\ 
 & (K=15) &46.55& 71.95&54.18 &58.19 &72.01 &55.28&\textbf{49.23} & \textbf{72.41}& \textbf{54.25}& \textbf{59.67}& \textbf{74.58}& \textbf{56.67} \\ \hline

\multirow{3}{*}{BP}
 & (K=~5~) &25.25&\textbf{59.52}&48.49&\textcolor[rgb]{0.45,0.45,0.45}{40.32}&55.73&48.68 &\textbf{26.87}&58.35&\textbf{52.54}&\textcolor[rgb]{0.45,0.45,0.45}{39.48}&62.52&\textbf{49.95} \\
 & (K=10) &27.54&61.48&51.11&\textcolor[rgb]{0.45,0.45,0.45}{44.33}&60.15&51.56 &\textbf{28.19}&\textbf{62.32}&\textbf{55.86}&\textcolor[rgb]{0.45,0.45,0.45}{43.25}&66.38&\textbf{52.92} \\
 & (K=15) &28.29&\textbf{63.98}&52.18&47.32&61.02&53.36 &\textbf{29.25}&63.15&\textbf{57.60}&\textcolor[rgb]{0.45,0.45,0.45}{43.26}&\textbf{68.83}&\textbf{54.16} \\ \hline
\end{tabular}
\end{table*}

Figure \ref{fig:analysis} shows the performance comparison between baseline models and our framework given different amount of training data on FewRel dataset. 
Each subgraph contains 6 groups, one group for each baseline model. For each group, the training data size increases from the left to right. 
Here, we focus on the blue curves which present baseline accuracies, the red curves which present the MICK enhanced model accuracies, and the gray bars which present the performance gains that MICK brings.
As is illustrated, performance of models deteriorates with the decrease of training data size. For example, the prototypical networks achieves 57.53\% accuracy with full training data under 10-way 1-shot test tasks, but performs poorly with 22.30\% accuracy given only 0.22\% of full training data.
Our framework fits in situations where only small training data is available. As is shown in Figure \ref{fig:analysis}, with our methods, the less the training data, the more improvement the model tends to gain. This indicates the effectiveness of our framework under extremely small training data. While our methods only improve prototypical networks by 1.27\% accuracy with full training data under 10-way 1-shot test tasks, it leads to 22.74\% improvement given 0.22\% of full training data.

Table \ref{FRMresult} shows the experimental results on the TinyRel-CM dataset.
Our framework considerably improves model performance in most cases.
Strong baselines on FewRel dataset such as Bert-Pair do not perform well, partially because the TinyRel-CM dataset is more verbal and informal, thus quite distinct from BERT-Chinese's pretraining data (Chinese Wikipedia).

On both datasets, when the input sentence contains multiple relation classes, model performance drops. E.g., under 2.23\% full FewRel training data under 5-way 5-shot tasks using MLMAN, classification accuracy on relation class \emph{mother} decreases by 14.14\% when \emph{spouse} or \emph{child} appears as interference.

\subsubsection{With sufficient training data}

We use full training data from FewRel dataset and apply MICK framework on each baseline model.

Here, we focus on the rightmost blue points, red points and gray bars of each group in \figref{fig:analysis}, which present the baseline performance, MICK performance and performance gains brought by MICK under full FewRel training data. As is shown in \figref{fig:analysis}, on full FewRel dataset, in most cases, our framework achieves better performance than baseline models.
The framework brings more improvement for relatively poor baselines such as GNN and SNAIL (about 5\% accuracy) than strong baselines such as MLMAN and Bert-Pair (about 1\% accuracy). This is because our framework aims to help models learn better and doesn't change the core part of the models. Full training data is sufficient for strong baselines to train a good model, leaving limited room for improvement. 


\subsubsection{Ablation tests}

Here, we focus on the individual effects of support classifier and task enrichment on baseline models. 

As is shown in Figure \ref{fig:analysis} and Table \ref{FRMresult}, in most cases, either adding support classifier or task enrichment improves performance for baseline models.
This illustrates that both support classifier and task enrichment contribute to better performance.

On FewRel dataset, we focus on the green and orange curves in \figref{fig:analysis}, which present the accuracies of applying support classifier or task enrichment individually.
Support classifier generally brings more performance gains when less training data is given. This is because with small training data, baseline models fail to extract adequate knowledge while the support classifier helps with the knowledge learning process.
In some rare cases such as SNAIL under 5-way 1-shot and 5-way 5-shot scenarios with 0.22\% training data, adding support classifier leads to not much improvement. 
This is because support classifier guides models to extract more knowledge from very limited resources, while the extractable knowledge is restricted by the training set size and some of the knowledge is even useless.
Support classifier improves poor baselines such as GNN and SNAIL to a larger extent than strong ones such as MLMAN and Bert-Pair because the support classifier makes up for the insufficient learning ability for poor models while is icing on the cake for strong models. 

Task enrichment is much more helpful than support classifier under small training data (when training data is shrunken to less than 1.00\%). This is because task enrichment compensates for the lack of learning sources of basic knowledge such as basic rules and grammar when training data is quite limited.
Improvement of task enrichment is also more obvious on poor baselines than strong ones. Since poor models fail to master some of the basic knowledge brought in original training data, the introduction of cross-domain data not only brings extra basic knowledge but also helps models to master knowledge in original training data by providing more related data.

Applying both support classifier and task enrichment achieves best performance in most cases, because the two methods complement each other. Support classifier extract useful information from original training data and cross-domain data, and task enrichment provide extra sources for the support classifier. 
In some cases, although adding either support classifier or task enrichment does not affect much, applying them together leads to a great improvement (e.g., HATT under 0.22\% training data).


On TinyRel-CM dataset (\tabref{FRMresult}), 
task enrichment brings about similar improvements for all baselines due to small training data. Support classifier leads to more performance gains than task enrichment individually,
indicating TinyRel-CM dataset is more challenging (the relation classes are much more similar than FewRel dataset) and baseline models fail to extract sufficient useful knowledge.
Although task enrichment brings not much improvement or even negative effects in some cases, adding support classifier simultaneously tend to raise the accuracy to a large extent (up to 10\%). 
The reason is that with insufficient learning ability 
(although baselines such as MLMAN and Bert-Pair are relatively strong, their learning ability is still insufficient with small training data),
cross-domain data alone sometimes introduces
noise and thus confuses the model. But with the additional support classifier
that improves learning ability, models are able to learn more useful
knowledge from cross-domain data.

On 
TinyRel-CM dataset (\tabref{FRMresult}), few cases occur where adding both support classifier and task enrichment performs worse than adding only one of them
(e.g., Bert-Pair under group D-U, and prototypical networks under group D-S in TinyRel-CM dataset).
The main reason is that while adding the support classifier helps models learn more and better, the model occasionally tend to pay excessive attention to the distinctive distribution of the cross-domain data.
As is shown in the corresponding results, in the vast majority of cases, while adding support classifier on baseline models raises accuracy, adding support classifier on task enriched models brings less improvement or even bad effect. This indicates stronger learning ability owing to the support classifier, and the distraction caused by cross-domain data.

\begin{table}[ht]
	\centering
	\small
	\caption{Human evaluation result on TinyRel-CM dataset under 5-shot scenario.}
	\label{human}
	\begin{tabular}{|c|cccccc|}
		\hline
		\textbf{Group} & D-D & D-S & D-F & D-U & F-U & S-F \\
		\textbf{(N)} & (5) & (2) & (6) & (6) & (2) & (6) \\ \hline
		\textbf{Acc(\%)} & 92.65 &96.53 & 86.93 & 91.77 & 96.71& 84.97\\ \hline
	\end{tabular}
\end{table}

\subsubsection{Dataset analysis}
Our TinyRel-CM dataset is a challenging task.
We human evaluate the dataset and results are illustrated in Table \ref{human}.
During the human-evaluation process, under $N$-way $K$-shot scenario, we provide instances of $N$ relation classes with $K$ instances per relation class. We use labels $1$ to $N$ to name the relation classes instead of their real names. A person is required to classify a new coming instance into one of the $N$ classes. 
We only evaluate the TinyRel-CM dataset under 5-shot scenarios because 10-shot and 15-shot tasks are too easy for human.
Three volunteers participated in the human evaluation process and we take the average accuracy as the final result.

Comparing Table \ref{FRMresult} and Table \ref{human}, the performance of state-of-the-art models is still far worse than human performance, indicating the TinyRel-CM dataset is a challenging task.

\section{Related Work}
Relation classification task aims to categorize the semantic relation between two entities conveyed by a given sentence into a relation class.
In recent years, deep learning has become a major method for relation classification.
Zeng et al. \shortcite{zeng-etal-2014-relation} utilized a convolutional deep neural network (DNN) during relation classification to extract lexical and sentence-level features. Vu et al. \shortcite{vu-etal-2016-combining} employed a voting scheme by aggregating a convolutional neural network with a recurrent neural network.
Traditional relation classification models suffer from lack of data. To eliminate this deficiency, distant-supervised approaches are proposed, which take advantage of knowledge bases to supervise the auto-annotation on massive raw data to form large datasets. Riedel et al. \shortcite{NYTdataset} constructed the NYT-10 dataset with Freebase \cite{Freebase} as the distant supervision knowledge base and sentences in the New York Times (2005-2007) as raw text. Distant-supervised methods suffer from long tail problems and excessive noise. Zeng et al. \shortcite{zeng-etal-2015-distant} proposed piecewise convolutional neural networks (PCNNs) with instance-level attention to eliminate the negative effect caused by the wrongly labeled instances on NYT-10 dataset. Liu et al. \shortcite{liu-etal-2017-soft} further introduced soft label mechanism to automatically correct not only the wrongly labeled instances but also the original noise from the distant supervision knowledge base.

The lack of annotated data leads to the emergence of few-shot learning,
where models need to perform classification tasks without seeing much data.
Meta-learning is a popular method for few-shot learning and is widely investigated in computer vision (CV).
Lake et al. \shortcite{LakeHuman} proposed Omniglot, a few-shot image classification dataset and put forward the idea of \emph{learning to learn}, which is the essence of meta-learning.
Memory-Augmented Neural Networks \cite{Santoro2016} utilized a recurrent neural network with augmented memory to store information for the instances the model has encountered.
Meta Networks \cite{metanet} implemented a high-level meta learner based on the conventional learner to control the update steps of the conventional learner.
GNN \cite{gnn} regarded support and query instances as nodes in a graph where information propagates among nodes, and classified a query node with the information of support nodes.
SNAIL \cite{snail} aggregated attention into the meta learner.
Prototypical networks \cite{proto} assumed that each relation has a prototype and classified a query instance into the relation of the closest prototype.
Image deformation meta-networks \cite{chen-2019-image} utilized a image deformation sub-net to generate more training instances for one-shot image classification.

Few-shot relation classification is a newly-born task that requires models to do relation classification under merely a few support instances.
Han et al. \shortcite{han-etal-2018-fewrel} proposed the FewRel dataset for few-shot relation classification and applied meta-learning methods intended for CV, including Meta Networks \cite{metanet}, GNN \cite{gnn}, SNAIL \cite{snail}, and prototypical networks \cite{proto}, 
on the FewRel dataset.
Prototypical networks with CNN core turned out to have the best test accuracy among the reported results.
Models that are more applicable for relation classification tasks are further proposed.
ProtoHATT \cite{hatt} reinforced the prototypical networks with hybrid attention mechanism. MLMAN \cite{ye-ling-2019-multi} improved the prototypical networks by adding mutual information between support instances and query instances. Bert-Pair \cite{gao-etal-2019-fewrel} adopted BERT \cite{devlin2018bert} to conduct binary relation classifications between a query instance and each support instance and fine-tuned on FewRel dataset. Baldini Soares et al. \shortcite{baldini-soares-etal-2019-matching} trained a BERT-like language model on huge open-source data with a Matching the Blanks task and applied the trained model on few-shot relation classification tasks.

Meta networks perform worst among all the methods \cite{han-etal-2018-fewrel} and
is time consuming (about 10 times slower than other methods with even better results).
Matching the Blanks is high-resource and not comparable to other methods. So we do not adopt these two methods as baselines.

Previous methods lose sight of the significant knowledge within the support instances.
Chen et al. \shortcite{chen-2019-image} conducted one-shot image classification task with an image deformation sub-net. The sub-net is designed to generate more support instances to augment labeled data and is trained by a prototype classifier. The prototype classifier updates the deformation sub-net according to the classification results on prototypes of generated support instances, and aims to improve the deformation process.
Inspired by this work, we add a support classifier over each support instance during the training process. The support classifier helps with the update process of the parameters in both the support classifier and the encoder, and is scheduled with a fast-slow learner scheme. The support classifier aims to utilize knowledge within support instances to obtain better instance representations. Additionally, previous few-shot relation classification models are trained with sufficient training data despite the small support set size during testing, 
we put forward a new challenge on few-shot relation classification by limiting the training data size.

Relation classification datasets have been released in past decades. Conventional ones include SemEval-2010 Task 8 dataset \cite{semeval8}, ACE 2003-2004 dataset \cite{ace}, TACRED dataset \cite{zhang-etal-2017-position} and NYT-10 dataset \cite{NYTdataset}. All these datasets encompass sufficient data to train a strong model.
Han et al. \shortcite{han-etal-2018-fewrel} released the first few-shot relation classification dataset, the FewRel dataset, which contains 100 relation classes with 700 instances per class. 
Although only few support instances are provided in each test task, the training data is sufficiently large. 
We propose TinyRel-CM dataset, 
a brand new few-shot relation classification dataset with purposely small training data and challenging relation classes.
The TinyRel-CM dataset is the second and first Chinese few-shot relation classification dataset.

\section{Conclusion}
In this paper, we propose a few-shot learning framework for relation classification that aims to (1) extract intra-support knowledge by classifying both support and query instances, and (2) bring external implicit knowledge from cross-domain corpus by task enrichment.
Our framework is particularly powerful when only small amount of training data is available. 
Additionally, we construct our own dataset, the TinyRel-CM dataset, a Chinese few-shot relation classification dataset in medical domain. The small training data size and highly similar 
relation classes make the TinyRel-CM dataset a challenging task.
As for future work, we intend to futher investigate whether the proposed framework is able to handle zero-shot learning tasks.

\section*{Acknowledgement}
This work was partially supported by NSFC grant 91646205, SJTU Medicine-Engineering Cross-disciplinary Research Scheme and SJTU-Leyan Joint Research Scheme.

\end{CJK}

\bibliographystyle{ACM-Reference-Format}
\bibliography{ref}

\end{document}